\documentclass{IMAGE2025}

\usepackage{amsmath, hyperref}
\begin{document}

\title{OpenSeisML: Open Large-Scale Real Seismic and well-log Dataset for Generative AI}

\renewcommand{\thefootnote}{\fnsymbol{footnote}} 

\author{Ipsita Bhar$^1$, Huseyin Tuna Erdinc$^1$, Thales Souza$^1$, Charles Jones$^2$ and Felix J. Herrmann$^1$\newline
$^1$Georgia Institute of Technology, Atlanta, USA \quad $^2$Osokey Ltd, Henley-on-Thames, UK}

\maketitle

\begin{abstract}
The advent of machine learning (ML) and computer vision has significantly accelerated seismic inversion workflows by reducing the computational cost of traditionally expensive iterative methods. However, the development and evaluation of ML methods remain limited by the scarcity of realistic velocity models, as most high-quality data are privately owned by oil and gas companies. To address this gap, we present \textit{OpenSeisML}, a collection of real seismic datasets designed to support generative AI (Gen-AI) workflows for seismic inversion. The datasets are curated from publicly available surveys in the UK National Data Repository (NDR). When seismic volumes are in the time domain and wells are in depth, a time-to-depth conversion is required. We use checkshot data to establish the time–depth relationship and construct a velocity model through interpolation for accurate conversion of post-stack seismic data. Here, we present an automated data curation pipeline that enables seismic data preparation while ensuring reproducibility.
The objective is to train a generative model that captures the statistical distribution of subsurface properties, enabling the synthesis of multiple statistically consistent realizations for uncertainty quantification which can act as a prior for seismic inversion.
\end{abstract}

\vspace{-4mm}
\section{Introduction and Related Work}
\vspace{-3mm}
Recent advances in machine learning show that progress depends not only on model improvements, but also on standardized benchmarks, shared datasets, and reproducible workflows that enable fair comparison of different ML models under consistent data, metrics, and protocols, ensuring reproducible and trustworthy results \citep{Donoho2024Data}. 
The lack of publicly accessible high-quality seismic datasets, especially realistic velocity models, limits the development and validation of Gen-AI workflows, as also emphasized by \citep{jin2024largescale} and \citep{alaudah2019facies}. To mitigate this bottleneck, we present an automated seismic data curation pipeline that provides large scale real datasets consisting of imaged seismic volumes and well-log data. In addition to velocity related information, the well dataset also includes petrophysical measurements such as gamma-ray, neutron, density, sonic, and resistivity logs, which provide complementary constraints on lithology, porosity, and fluid properties.

Synthetic datasets have played a major role in developing new algorithms to train ML models for a myriad of subsurface applications such as underground energy storage \citep{gahlot2025digitalshadow}; \citep{hydro}, reservoir characterization \citep{chen2025diffusion}, etc. These models rely on large-scale good quality training datasets to achieve reliable inference. Some of these synthetic datasets are discussed below.


\noindent\textbf{Compass model and SEAM open data}: The Compass model was constructed to evaluate acquisition strategies and velocity inversion methods \citep{yin2024wise}; \citep{yin2025wiser}; \citep{orozco2025aspire}, by mimicking real geological complexity by incorporating features such as faults, folds, channels, and gas clouds derived from real seismic and well-log data from the North Sea \citep{jones2012bgmodel}. The model was built through a manual workflow by integrating well logs, horizons, and seismic attributes. Although realistic, the authors constructed only a single subsurface realization, thus limiting variability, so a generative model trained on it may overfit to that specific geology and struggle to capture the broader distribution of subsurface structures across different regions \citep{jin2024largescale}. SEAM \citep{seam} and \citep{pangman2014seam}, is another realistic subsurface benchmark, developed for research advancement in seismic imaging and inversion . They incorporated realistic geological features such as salt bodies, stratigraphy, and structural complexity based on field data, however, the simulations still remain synthetic and controlled. This dataset also represents a limited number of deterministic realizations, whereas ML methods require diverse datasets with multiple realizations to adequately capture subsurface variability.

\noindent\textbf{OpenFWI Benchmarks}: 
To support the development and comparison of data-driven inversion methods, the OpenFWI benchmark provides a large-scale, open-access collection of synthetic seismic datasets \citep{deng2022openfwi}. These were created by generating large-scale synthetic datasets using numerical wave equation solvers with diverse velocity models. The authors designed multiple geological scenarios with varying structures and acquisition settings to produce paired seismic data and ground-truth models. This controlled setup enables supervised training and systematic evaluation of data-driven inversion methods; however, these datasets do not fully capture the complexity and variability of real field data \citep{jin2024largescale}. As a result, models trained on these datasets tend to learn mappings specific to the synthetic setup and may struggle to generalize to more complex or realistic subsurface conditions \citep{consolvo2023openfwi}.

\begin{figure*}[h]
\centering
\includegraphics[width=0.8\linewidth]{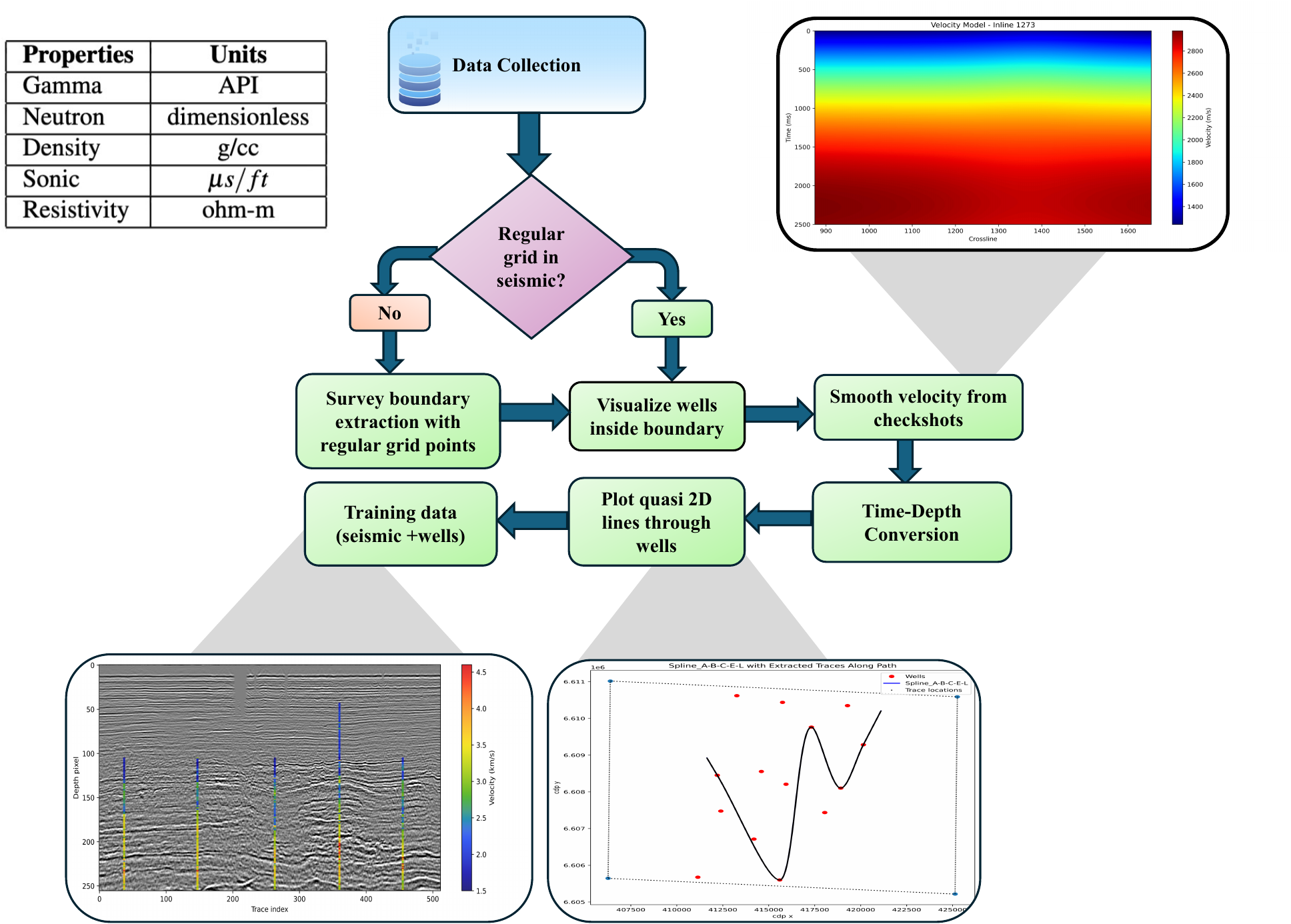}
\caption{\footnotesize The flow diagram for data curation pipeline and the table shows the well logs along with their units present in the las files.}
\label{fig:depth}
\end{figure*}

\section{Our Contribution}
\vspace{-3mm}
From the above discussion, it is evident that while some models are geologically realistic, they are limited to single realizations and fail to capture stochastic variability, whereas others, despite their scale, lack sufficient geological complexity to represent real 
subsurface conditions. Hence, we wanted to address this challenge by curating a training set based on real-field seismic data from the UK National Data Repository (NDR) \citep{ukndr_nsta}, a government-supported platform structured access to subsurface data across the UK energy 
sector. The curated dataset is intended to support the training of generative models that produce 
statistically consistent and geologically plausible subsurface realizations representative of the 
North Sea basin, while maintaining the ability to generalize across diverse marine environments. 
We have automated our data curation pipeline to produce structured and consistent datasets 
that can be used directly for training without additional preprocessing.

\vspace{-3mm}
\section*{Machine Learning Data Curation Pipeline}
\vspace{-3mm}
The data curation workflow, illustrated in {Figure 1}, consists of four components, which are described below.


\subsection{Data Collection}

The UK-NDR portal enables bulk download of 
3-D migrated seismic data, as illustrated in {Figure 2}. For quality control, we prioritized surveys acquired after the 1990s, as earlier datasets often contain inconsistent SEG-Y headers, nonstandard formatting, and uncertain well coordinates that complicate automated geometry extraction and well–seismic alignment. A small number of pre-1990 surveys were included only after careful verification. For each selected survey, we downloaded SEG-Y volumes directly, and retrieved associated well data (LAS and checkshot files) using automated scripts that access the repository through APIs. This provides a scalable and reproducible workflow for downloading large seismic and well datasets.

\begin{figure}[h]
\centering
\includegraphics[width=0.8\linewidth]{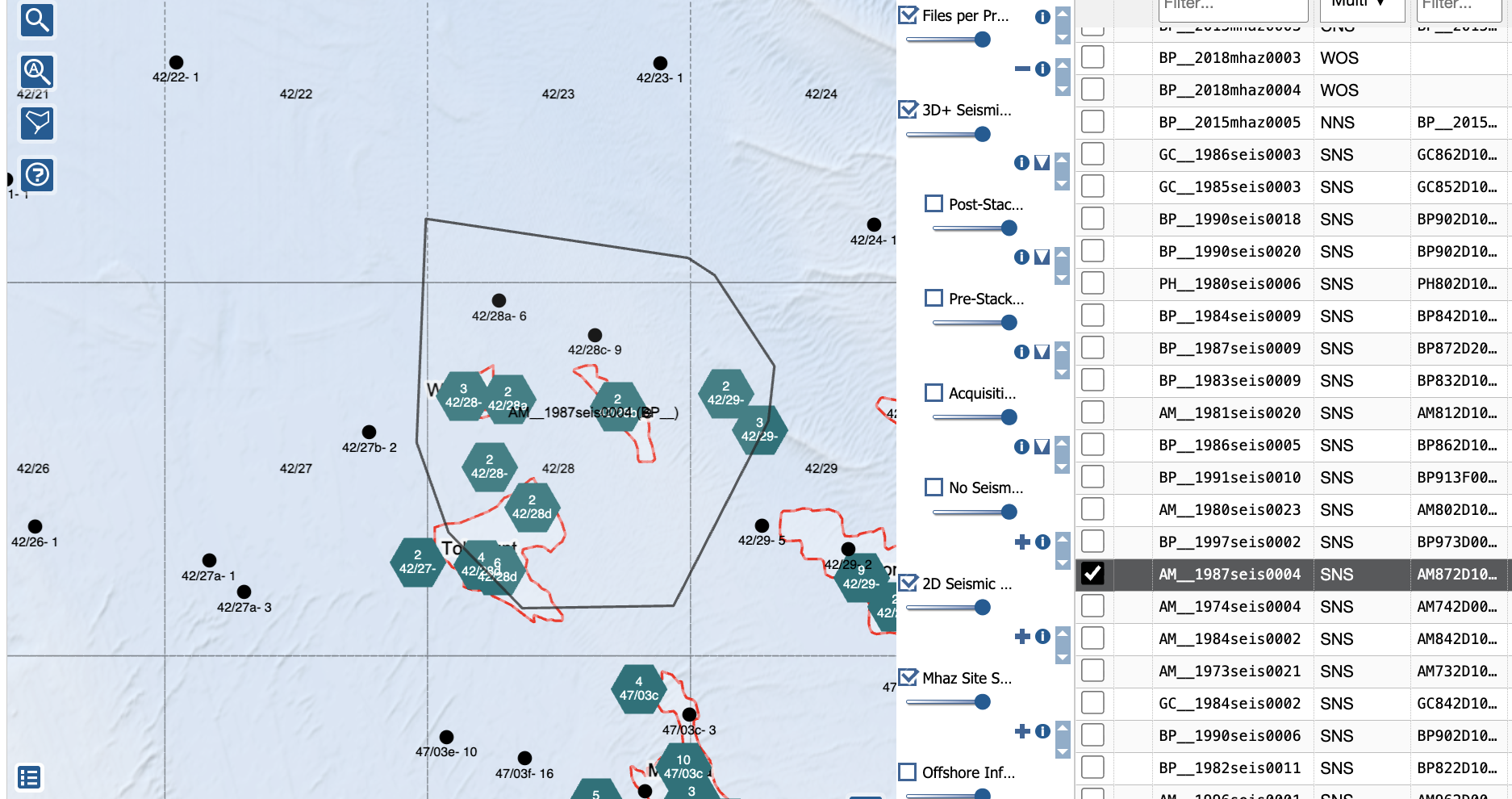}
\caption{\footnotesize UKNDR GUI for seismic data filtering and downloading}
\label{fig:depth}
\end{figure}

\subsection{Preprocessing of SEG-Y
Data}\label{preprocessing-of-seg-y-data}

After collecting the 3D seismic datasets, preprocessing is required to make the data suitable for generative model training. The first step involves aligning seismic volumes and well data within a common coordinate reference system (CRS) \citep{crsjanssen}, ensuring that wells are correctly positioned within their corresponding survey boundaries. To represent 3D seismic data on a regular grid, we first estimate survey boundaries using a concave hull to account for irregular acquisition geometries, and then extract the largest contiguous rectangular region with regular grids \citep{gfg_histogram}.

\subsection{Checkshot-Based Velocity Volume Construction for Time-Depth Seismic Conversion}

Post-stack seismic volumes are migrated data and may be available in either the time or depth domain. In cases where the data are time migrated, a time-to-depth conversion is required for integration with well information. Hence, we propose to use checkshot data, which provide depth and corresponding two-way travel times, to compute average velocities for time-depth conversion of time migrated seismic volumes. These velocity estimates are then spatially
interpolated using a Radial Basis function (RBF) \citep{skala2017rbf} to construct a three dimensional average velocity volume as shown in {Figure 3}.

The general RBF interpolant is defined as: 
\begin{equation}
f(x) = \sum_{i=1}^{N} \lambda_i \, \phi(\|x - x_i\|)
\end{equation}
\begin{itemize}
    \item $x$ represents a spatial location where velocity is estimated (e.g., grid point in the seismic volume: $(x,y)$ or $(x,y,z)$)
    \item $x_i$ are the locations of known data points (checkshot positions)
    \item $f(x)$ represents interpolated value, i.e., velocity at location $x$
    \item $\lambda_i$ weights determined from known velocities at the wells, and
    \item $\phi(\|x - x_i\|)$ radial basis function that depends on the distance between $x$ and each data point $x_i$
\end{itemize}

We intentionally constructed an average velocity volume rather than an
interval velocity, in order to obtain a smooth velocity field
for depth conversion. Although interval velocity is an intrinsic rock property, our objective is not to construct highly accurate velocity models, but rather to obtain a consistent mapping between time and depth that enables alignment of seismic data with well information. In practice, estimating precise interval velocities is challenging \citep{kosloff2002interval}, particularly in areas with sparse well control or limited data quality. Velocity information is typically derived from time–velocity measurements such as stacking velocities or checkshot data \citep{herron2011seismic}.Instead of focusing on exact velocity reconstruction, we aim to capture the overall statistical behavior of the velocity field. 


As part of quality control for the interpolated velocity model, we performed consistency checks between the original checkshot measurements and the interpolated velocity values at corresponding well locations shown in {Figures 4 (a) and 4 (b)}, comparing time–velocity and depth–velocity profiles derived from checkshots with those obtained from the interpolated average velocity volume at the same spatial positions. The depth versus velocity and two way travel time vs velocity trends show good agreement, indicating that the interpolated velocity model preserves the underlying well control concluding that the constructed average velocity volume remains consistent with measured checkshot data and is suitable for reliable time–depth conversion. 

\begin{figure}
\centering
\includegraphics[width=0.8\linewidth]{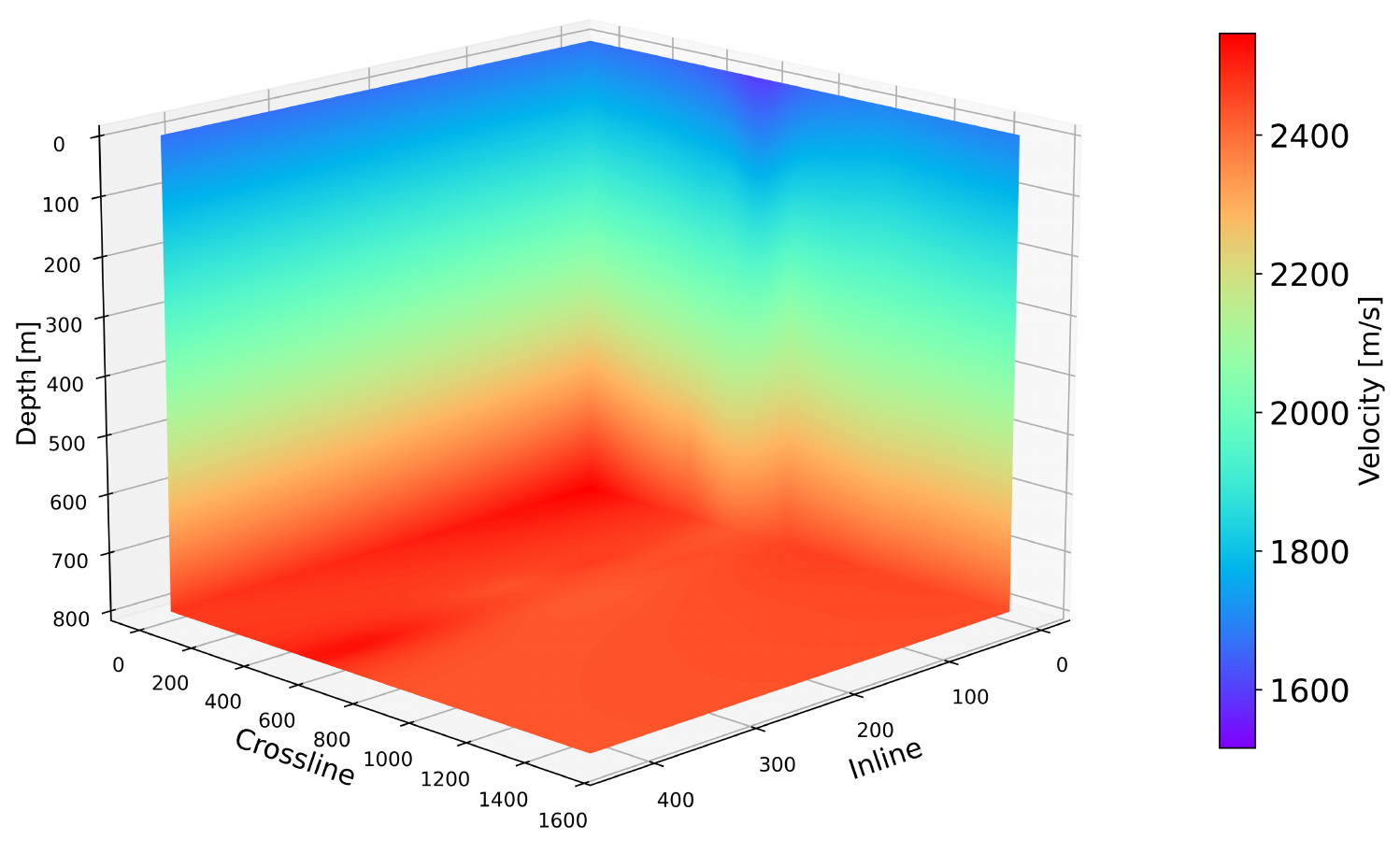}
\caption{\footnotesize 3-D visualization of smooth velocity field constructed using checkshots}
\label{fig:depth}
\vspace{-3mm}
\end{figure}

\begin{figure}[h]
\centering
\includegraphics[width=0.98\linewidth]{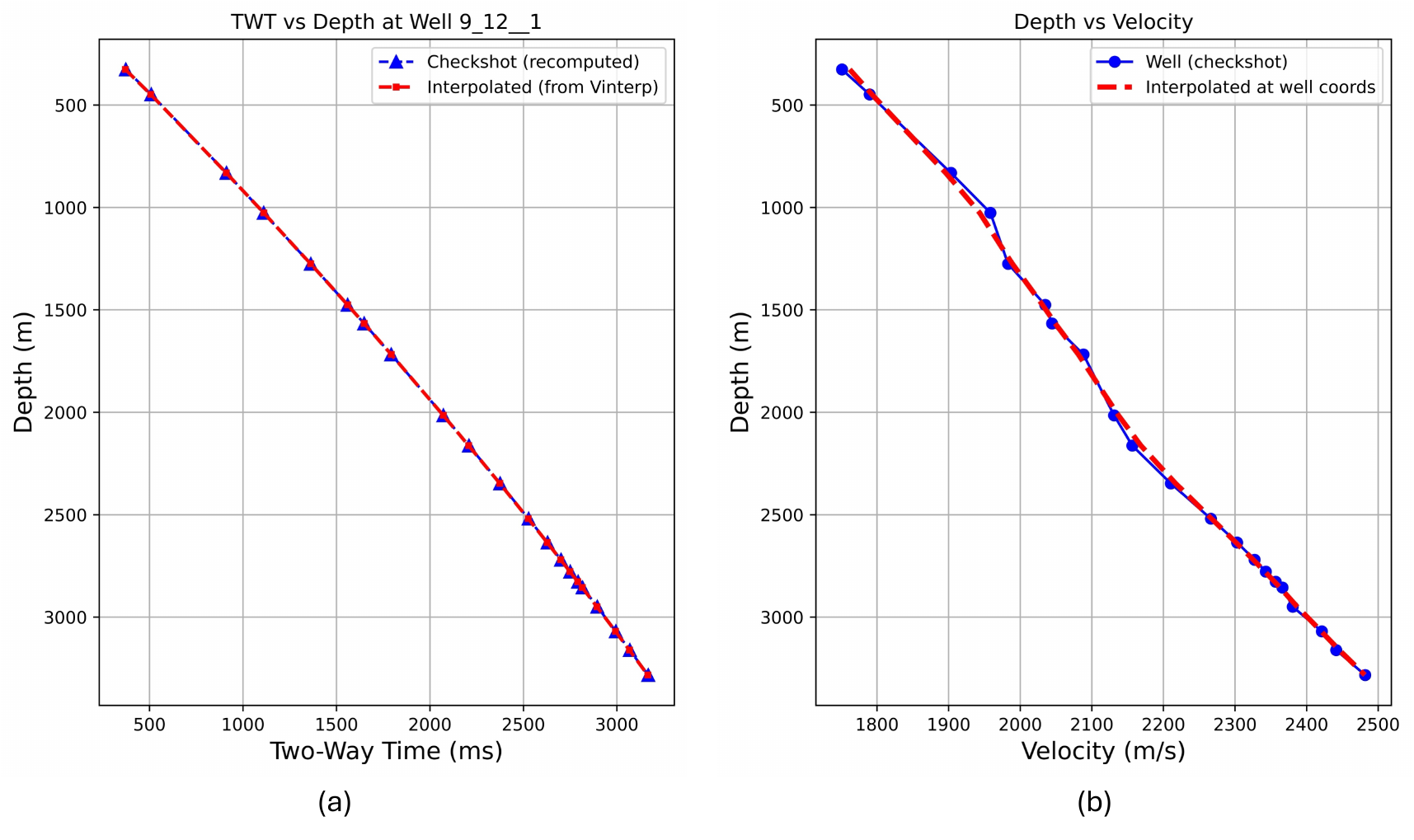}
\caption{\footnotesize(a) Comparison of two-way travel time versus depth for checkshot data (blue curve) and the corresponding interpolated well velocity at the same location (red curve) (b) Comparison of depth vs velocity for checkshot measurements and the corresponding interpolated well velocity at the same location}
\label{fig:depth}
\vspace{-3mm}
\end{figure}

\subsection{Time to Depth Conversion of Seismic
Data}\label{time-to-depth-conversion-of-seismic-data}

The time migrated seismic volumes are converted to depth in OpendTect \citep{OpendTect} and the depth converted seismic is illustrated in {Figure 5}. The
conversion from two way travel time to depth is governed by the
fundamental relationship between velocity and travel time.
For vertically propagating waves, depth is obtained by integrating
interval velocity over time: $z(t) = \int_{0}^{t} \frac{v(\tau)}{2} \, d\tau$ where, $z(t)$ denotes depth, $v(\tau)$ is the interval velocity as a function of time, and $t$ represents the two-way travel time.
The factor of \(\frac{1}{2}\) accounts for the fact that seismic data
are recorded in two way travel time. The depth at time sample \(t_k\) is computed as $z_k = \sum_{j=1}^{k} \frac{v_j}{2}{\times}\Delta{t}$.
Although, not accurate, but this limitation is acceptable as the objective is to learn the statistical distribution of velocity models that at a later stage can be used to train generative neural networks to carry out seismic inference.

\subsection{Training Data}\label{training-data-preparation}

The final step in the data curation pipeline is to generate two-dimensional depth-domain seismic sections from curated three-dimensional volumes. The well locations are first mapped within the seismic boundary, after which quasi-2D lines are constructed along the trajectories that pass through the selected wells as depicted in {Figure 5}.

\begin{figure}[h]
\centering
\includegraphics[width=0.98\linewidth]{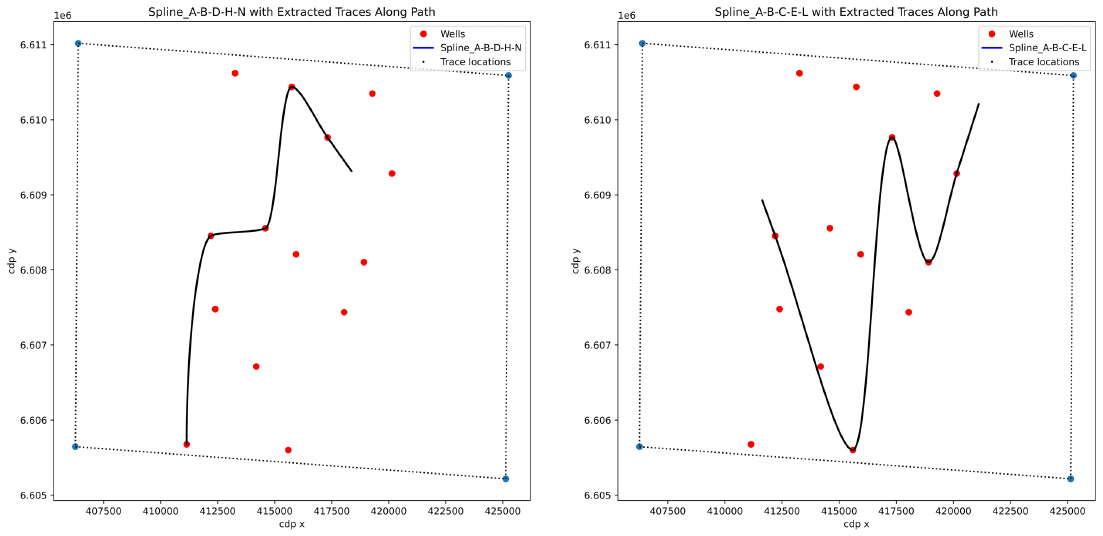}
\caption{\footnotesize Quasi-2D lines passing through different well locations}
\label{fig:depth}
\end{figure}



The extracted seismic sections corresponding to the quasi-2D lines were resampled to 256×512 using a 2D FFT, where a smooth low-pass filter with a cosine taper suppresses high-frequency components, preserving low frequencies while gradually attenuating frequencies to avoid sharp cutoffs. Since well logs typically have finer vertical sampling than seismic traces, filtering and resampling are applied to wells to harmonize resolutions with seismic before training of a diffusion based Gen-AI model. For details on this training, we refer to another abstract by the authors submitted to the proceedings of this conference. The final data set consists of 2-D seismic sections in depth-domain of dimension 256x512 with a 12.5m sampling interval integrated with well control, as illustrated in {Figure 7}. All curated datasets are stored in HDF5 format.

\begin{figure}[h]
\centering
\includegraphics[width=0.98\linewidth]{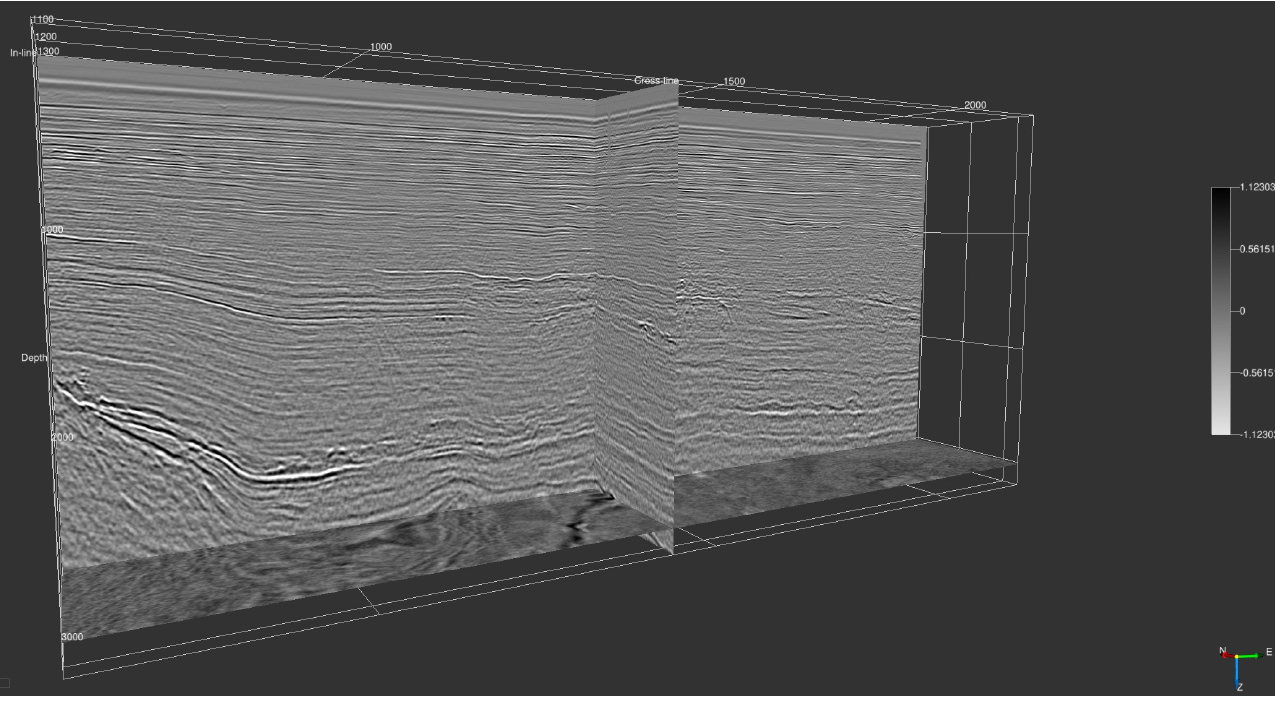}
\caption{\footnotesize Seismic image after converting from Time to Depth domain visualised in OpendTect}
\label{fig:depth}
\end{figure}

\begin{figure}[h!]
\centering
\includegraphics[width=0.98\linewidth]{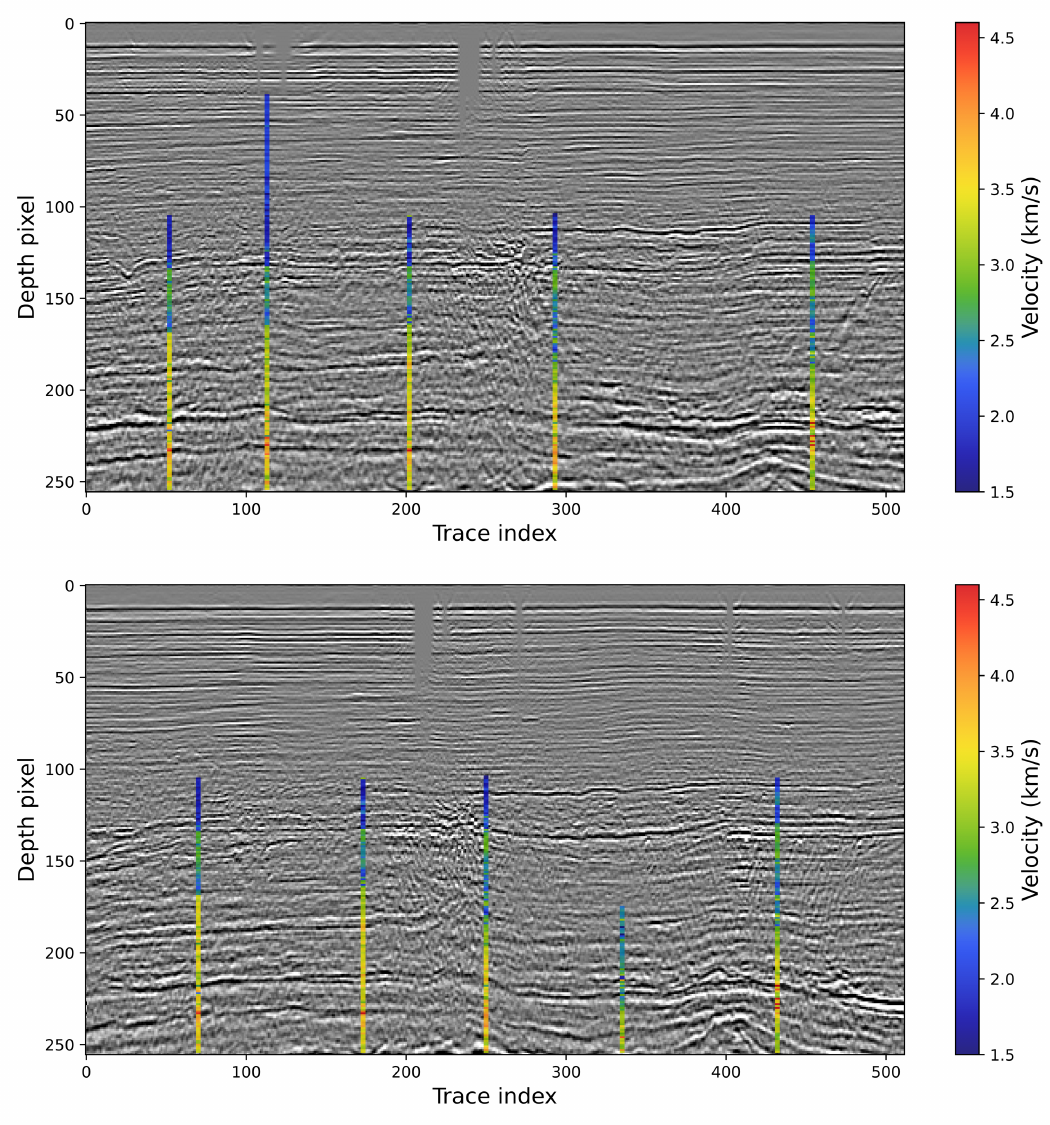}
\caption{\footnotesize Both the figures show different sections of 2-D resampled seismic in depth tied to wells}
\label{fig:depth}
\end{figure}

\newpage
\vspace{-3mm}
\section{RESULTS}
\vspace{-3mm}
A diffusion-based generative model \citep{erdinc2024generative} was initially trained on Compass dataset having the same dimension and sampling interval as these curated datasets. For the initial experiments, the model was retrained using our curated data as represented in {Figure 7}, from only 40 wells, and was able to generate realistic velocity model distributions that closely align with the ground truth as shown in {Figure 8}. These results are preliminary and future work will involve scaling the training to at least 1000 real well logs to further improve the model’s accuracy and generalization.

\begin{figure}
\centering
\includegraphics[width=0.98\linewidth]{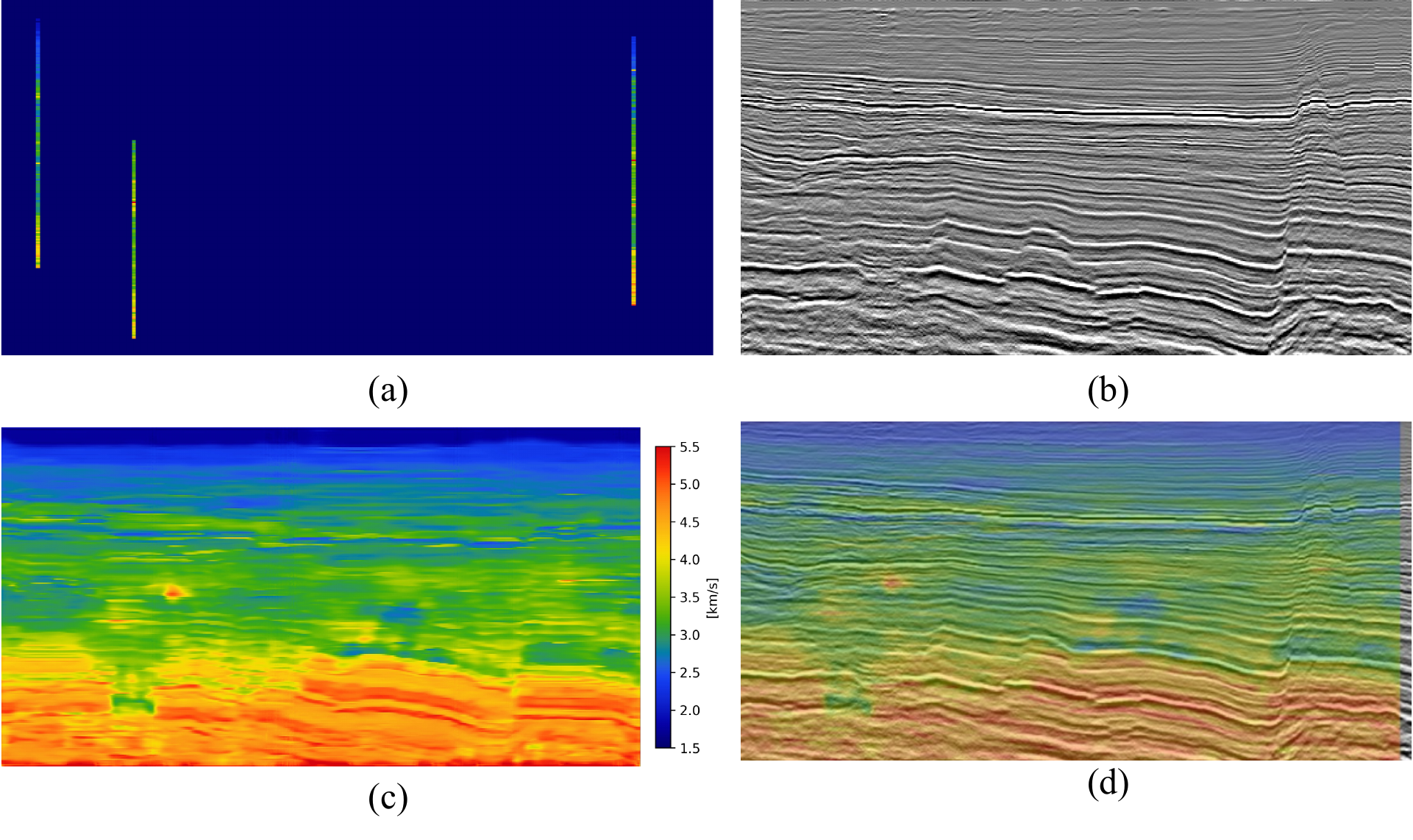}
\caption{\footnotesize (a) represents the ground truth velocity the network learns from the wells, (b) represents the corresponding seismic section, (c) is a velocity sample as reconstructed by the diffusion model, and (d) presents the overlaid velocity sample on the seismic image.}
\label{fig:depth}
\end{figure}


\vspace{-3mm}
\section{CONCLUSION}
\vspace{-3mm}
We introduce an automated data curation pipeline that streamlines seismic data preparation and by leveraging real field datasets, the proposed framework addresses key limitations of existing synthetic benchmark velocity models, including the lack of realistic geological complexity and variability. Unlike prior approaches that rely on deterministic models or simplified synthetic data, our methodology focuses on capturing the statistical characteristics from real seismic observations. These curated datasets include not only velocity but also additional properties such as density, enabling their use in evaluating multiparameter inversion algorithms in the future. Currently, we evaluate these datasets in a 2-D setting and are in the process of extending our framework to 3-D.

\subsection{ACKNOWLEDGMENTS}
This work was carried out in collaboration with the UK National Data
Repository (NDR). \textit{Contains information provided by the North Sea Transition Authority and/or other third parties.}
During the preparation of this work, the authors used ChatGPT for
language refinement and to improve readability. After using this
service, the authors reviewed and edited the content as needed and take
full responsibility for the content of the publication.



\subsection{DATA AND MATERIALS AVAILABILITY}
Data associated with this work are available \href{https://ndr.nstauthority.co.uk/}{here} and the curated datasets with codes will be made available upon acceptance.

\onecolumn

\twocolumn

\bibliographystyle{seg}  
\bibliography{Seismic}

\end{document}